\documentclass[]{spie}  
\usepackage[]{graphicx}
\usepackage{amssymb}

\usepackage{epstopdf}
\usepackage{hyperref}
\usepackage{caption}
\usepackage{subcaption}
\usepackage{graphicx}
\usepackage{amsmath}
\usepackage{algorithm}
\usepackage[noend]{algpseudocode}
\usepackage{amssymb}
\usepackage{latexsym}

\title{Classification of Time-Series Images Using Deep Convolutional Neural Networks} 

\author{Nima Hatami, Yann Gavet and Johan Debayle
\skiplinehalf
Ecole Nationale Superieure des Mines de Saint-Etienne,\\ 
SPIN/LGF CNRS UMR 5307, 158 cours Fauriel, 42023 Saint-Etienne, France
}

\authorinfo{Further author information:\\N. Hatami: E-mail: nima.hatami@emse.fr, Telephone: +33 477420220\\
Y. Gavet: E-mail: gavet@emse.fr, Telephone: +33 477420170\\
J. Debayle: E-mail: debayle@emse.fr, Telephone: +33 477420219}

\begin{document} 
  \maketitle 

\begin{abstract}
Convolutional Neural Networks (CNN) has achieved a great success in image recognition task by automatically learning a hierarchical feature representation from raw data. While the majority of Time-Series Classification (TSC) literature is focused on 1D signals, this paper uses Recurrence Plots (RP) to transform time-series into 2D texture images and then take advantage of the deep CNN classifier. 
Image representation of time-series introduces different feature types that are not available for 1D signals, and therefore TSC can be treated as texture image recognition task. CNN model also allows learning different levels of representations together with a classifier, jointly and automatically.
Therefore, using RP and CNN in a unified framework is expected to boost the recognition rate of TSC. 
Experimental results on the UCR time-series classification archive demonstrate competitive accuracy of the proposed approach, compared not only to the existing deep architectures, but also to the state-of-the art TSC algorithms.

\end{abstract}



\section{INTRODUCTION}
\label{sec:intro}  

A time-series is a sequence of data points (measurements) which has a natural temporal ordering.  Many important real-world pattern recognition tasks deal with time-series analysis. Biomedical signals (e.g. EEG and ECG), financial data (e.g. stock market and currency exchange rates), industrial devices (e.g. gas sensors and laser excitation), biometrics (e.g. voice, signature and gesture), video processing, music mining, forecasting and weather are examples of application domains with time-series nature \cite{Ref1,Ref2,Ref3}. The time-series analysis motivations and tasks are mainly divided into curve fitting, function approximation, prediction and forecasting, segmentation, classification and clustering categories. In a univariate time-series classification, $x^{n} \rightarrow y^{n}$ so that $n$-th series of length $l$: $x^{n} = (x^{n}_{1}, x^{n}_{2}, .. x^{n}_{l})$ is associated with a class label $y^{n} \in \{1,2,..,c\}$. It is worth noting that although this paper is mainly focused on time-series classification problem, the proposed method can be easily adapted to the other tasks such as clustering and anomaly detection.

The existing Time-Series Classification (TSC) methods may be categorized from different perspectives. Regarding the feature types, "frequency-domain" methods include spectral analysis and wavelet analysis; while "time-domain" methods include auto-correlation, auto-regression and cross-correlation analysis. Regarding the classification strategy, it can also be divided into "instance-based" and "feature-based" methods. The former measures similarity between any incoming test sample and the training set; and assigns a label to the most similar class (the euclidean distance based 1-Nearest Neighbor (1-NN) and Dynamic Time Wrapping (DTW) are two popular and widely used methods of this category \cite{Ref4,Ref5}. The latter first transforms the time-series into the new space and extract more discriminative and representative features in order to be used by a pattern classifier, which aims of the optimum classification boundaries \cite{Ref6,Ref7,Ref8}.

Recently, Deep Learning (DL also known as feature learning or representation learning) models have achieved a high recognition rate for computer vision \cite{Ref9, Ref10, Ref11} and speech recognition \cite{Ref12, Ref13}. The Convolutional Neural Networks (CNN) is one of the most popular DL models. Unlike the traditional "feature-based" classification framework, CNN does not require hand-crafted features. Both feature learning and classification parts are unified in one model and are learned jointly. Therefore, their performances are mutually enhanced. Multiple layers of different processing units (e.g. convolution, pooling, sigmoid/hyperbolic tangent squashing, rectifier and normalization) are responsible to learn (represent) a hierarchy of features from low-level to high-level.

This paper investigates the performance of Recurrence Plots (RP) \cite{Ref14} within the deep CNN model for TSC. RP provides a way to visualize the periodic nature of a trajectory through a phase space and enables us to investigate certain aspects of the m-dimensional phase space trajectory through a 2D representation. Because of the recent outstanding results by CNN on image recognition, we first encode time-series signals as 2D plots, and then treat TSC problem as texture recognition task. A CNN model with 2 hidden layers followed by a fully connected layer is used.

\section{Related Work}

This section briefly reviews the recent deep learning contributions on the TSC task. Application of deep learning on TSC has not been fully explored until recently. There are two main types of approaches, when it comes to application of CNN on TSC: some modify the traditional CNN architecture and use 1D time-series signals as an input, while some others first transform 1D signals into 2D matrices and then apply CNN, similar to the traditional CNN for image recognition. A time delay neural network (TDNN) model is adopted for EEG classification \cite{Ref15}. However, their model has only one single hidden layer and is not deep enough to learn hierarchical features. Lee et al. \cite{Ref16} explored convolutional deep belief network (CDBN) for audio classification. They applied this unsupervised feature learning on frequency domain rather than in time domain. A multi-channel CNN has been proposed to deal with multivariate time-series \cite{Ref17}. Each time series signal is fed into a separate CNN, and afterwards outputs of all CNNs are concatenated and fed into a fully connected MLP classifier. Instead of using the raw signal and learn features automatically, Dalto \cite{Ref18} feed CNN with variables post-processed using input variable selection (IVS) algorithm. In \cite{Ref19} audio signals are transformed to time-frequency domain and fed into CNN. In another similar work by the same authors, CNN is applied to speech recognition within the framework of hybrid NN-HMM model \cite{Ref20}. A multi-scale CNN is proposed for TSC in \cite{Ref21} . In order to extract features at different scales and frequencies, it transforms signals by down-sampling and smoothing operators, followed by a local convolution stage. A deep CNN is applied on multichannel time-series signals of human activities \cite{Ref22}. A sliding window strategy is adopted to put time-series segments into a collection of short pieces of signals. This way, a 2D representation of a 1D time-series signal is obtained and a CNN model applied on 2D matrices (treating them similar to images).

\begin{figure}
   \begin{center}
   \begin{tabular}{c}
   \includegraphics[height=4cm]{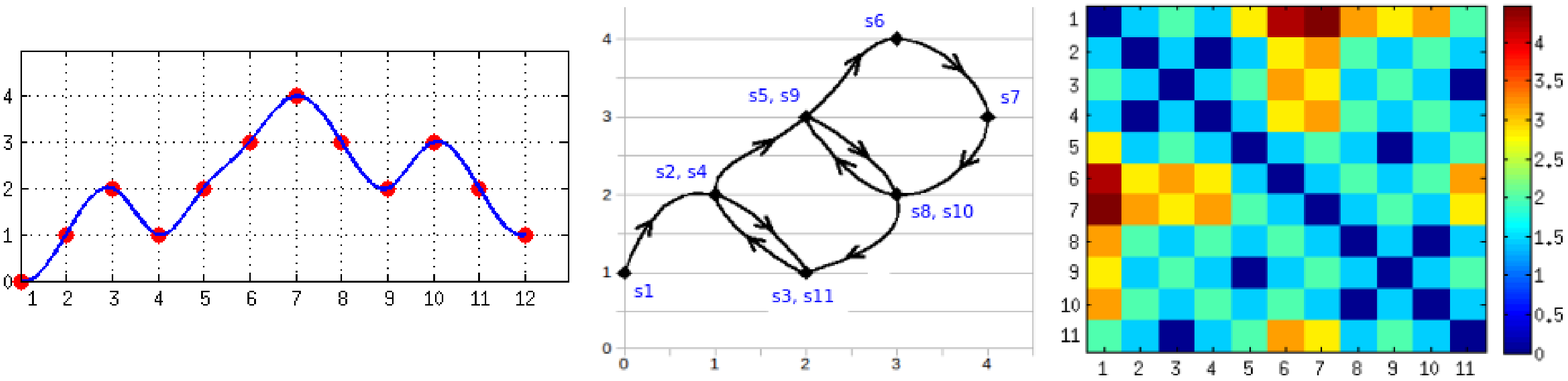}
   \end{tabular}
   \end{center}
   \caption{From time-series signal to recurrence plot. Left: A simple example of time-series signal ($x$) with 12 data points. Middle: The 2D phase space trajectory is constructed from $x$ by the time delay embedding ($\tau = 1$). States in the phase space are shown with bold dots: $s_{1}:(x_{1}, x_{2}), s_{2}:(x_{2}, x_{3}), ..., s_{11}:(x_{11}, x_{12})$. Right: The recurrence plot $R$ is a $11 \times 11$ square matrix with $R_{i,j}=dist(s_{i},s_{j})$.} 

  \label{fig:RP_toy} 
   \end{figure} 
   
Perhaps \cite{Ref23, Ref24} is the most similar research to our work here. The Gramian Angular Fields (GAF) and Markov Transition Fields (MTF) are used to encode time-series signals as images. Afterwards, a Tiled CNN model is applied to classify the time-series images. Another similar work applied visual descriptors such as Gabor and Local Binary Patterns (LBP) on RP to extract texture features from time series, and then used SVM classifiers \cite{Ref25, Ref26}. The difference between the latter and our work is that they used a traditional classification framework with hand-crafted features and separated feature extraction and classification steps. While our proposed CNN-based framework, automatically learns the texture features that are useful for the classification layer. The joint learning of feature representation and classifier offered by CNN increases the classification performance.   

\section{Methodology}
\subsection{Time-Series to Image Encoding}
Time-series can be characterized by a distinct recurrent behavior such as periodicities and irregular cyclicities. Additionally, the recurrence of states is a typical phenomenon for dynamic nonlinear systems or stochastic processes that time-series are generated in. The RP \cite{Ref14} is a visualization tool that aims to explore the m-dimensional phase space trajectory through a 2D representation of its recurrences. The main idea is to reveal in which points some trajectories return to a previous state and it can be formulated as:

\begin{equation}
\label{eq1}
R_{\rm i,j}=\theta(\epsilon-\Vert \vec{s}_{i} - \vec{s}_{j} \Vert), \hspace{2mm} \vec{s}(.) \in \mathfrak R^{m},  \hspace{2mm} i,j=1,...,K
\end{equation}

where $K$ is the number of considered states $\vec{s}$, $\epsilon$ is a threshold distance, $\Vert . \Vert$ a norm and $\theta(.)$ the Heaviside function. The R-matrix contains both \emph{texture} which are single dots, diagonal lines as well as vertical and horizontal lines; and \emph{typology} information which are characterized as homogeneous, periodic, drift and disrupted. For instance, fading to the upper left and lower right corners means that the process contains a trend or drift; or vertical and horizontal lines/clusters show that some states do not change or change slowly for some time and this can be interpreted as laminar states \cite{Ref27}. Obviously, there are patterns and information in RP that are not always very easy to visually see and interpret. 

Figure \ref{fig:RP_toy} shows the step-by-step instructions for calculating RP for a simple time-series signal. First the 2D phase space trajectory ($m=2$) is constructed from the time-series. Then, the R-matrix is calculated based on the closeness of the states in the phase space. It is worth noting that the resulting R-matrix by Formula \ref{eq1} has only \{0,1\} values, that caused by thresholding parameter $\epsilon$. In order to avoid information loss by binarization of the R-matrix, this paper skips the thresholding step and uses the gray-level texture images. Inspired by the unique texture images obtained from the R-matrices, this paper proposed a TSC pipeline based on the CNN model. First the raw 1D time-series signals $x^{n}$ are transformed into 2D recurrence texture images, and then both features and classifier are jointly learned in one unified model. 

   \begin{figure}
   \begin{center}
   \begin{tabular}{c}
   \includegraphics[height=3.5cm]{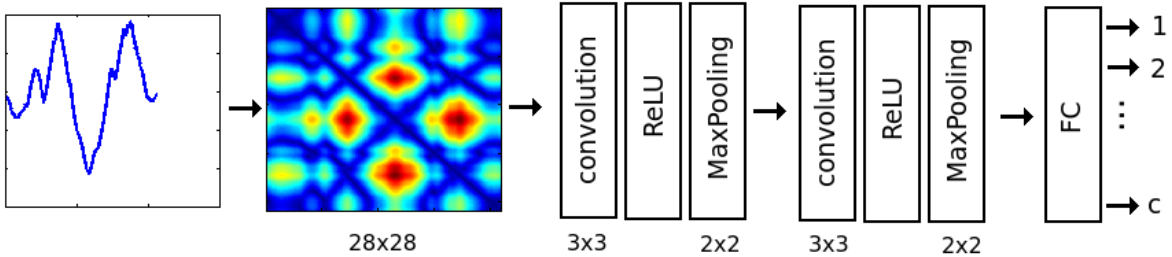}
   \end{tabular}
   \end{center}
   \caption{The proposed 2-stages CNN architecture for TSC. RP images are resized to 28x28, 56x56 or 64x64 (depend on the data) and fed into CNN model. This architecture \emph{32(5)-2-32(5)-2-125-c} consists of 2 convolution, 2 pooling, and 2 fully-connected layers.} 

  \label{fig:CNN} 
   \end{figure}

\subsection{CNN for Time-Series Image Classification}

There are two aspects of any CNN model that should be considered carefully: i) designing an appropriate architecture, and ii) choosing the right learning algorithm. Both architecture and learning rules should be chosen in a way that they are not only compatible with each other, but also fit the data and the application appropriately.

\textbf{Architecture.} A 2-stage deep CNN model is applied here with 1-channel input of size $28\times 28$ and the output layer with $c$ neurons. Each feature learning stage is representing a different \emph{feature-level} and consists of convolution (filter), activation, and pooling operators, respectively. The input and output of each layer are called \emph{feature maps}. A filter layer convolves its input with a set of trainable kernels. The convolutional layer is the core building block of a CNN and exploits spatially local correlation by enforcing a local connectivity pattern between neurons of adjacent layers. The connections are local, but always extend along the entire depth of the input volume in order to produce the strongest response to a spatially local input pattern. The activation function (such as \emph{sigmoid} and \emph{tanh}) introduces non-linearity into the networks and allows them to learn complex models. Here we applied ReLU (Rectified Linear Units) because it trains the neural networks several times faster \cite{Ref38} without a significant penalty to generalisation accuracy. Pooling (a.k.a. subsampling) reduces the resolution of input and make it robust to small variations for previously learned features. It combines the outputs of \emph{i-1}th layer into a single input in \emph{i}th layer over a range of local neighborhood.

At the end of the 2-stage feature extraction, the feature maps are flatten and fed into a fully connected (FC) layer for classification. FC layers connect every neuron in one layer to every neuron in another layer, which in principle are the same as the traditional multi-layer perceptron (MLP). The proposed pipeline for TSC is shown in Fig. \ref{fig:CNN}.\\


\begin{figure}
   \begin{center}
   \begin{tabular}{c}
   \includegraphics[height=5cm]{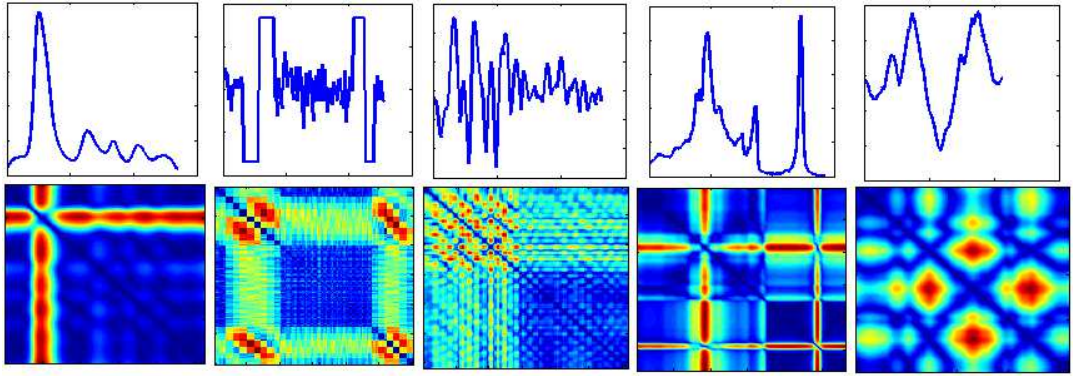}
   \end{tabular}
   \end{center}
   \caption{Application of RP ($m=3, \tau = 4$) time-series to image encoding on five different datasets from the UCR archive: 50words, TwoPatterns, FaceAll, OliveOil and Yoga data (from left to right, respectively).} 

  \label{fig:RP_UCR} 
   \end{figure} 
   
\textbf{Learning.} Training the above CNN architecture is similar to the MLPs. Gradient-based optimization method (error back-propagation algorithm) is utilized to estimate parameters of the model. For faster convergence, the stochastic gradient descent (SGD) is used for updating the parameters. The training phase has two main steps: propagation and weight update. Each propagation involves feedforward and error back-propagation passes. Former determines the feature maps on input vector by passing from layer to layer until reaching the output (left to right in Figure \ref{fig:CNN}). Latter, calculates the propagation errors according to the loss function for the predicted output (error propagates from right to left in Figure \ref{fig:CNN}). Predicted error on each layer is used for calculating the derivatives by taking advantage of chain-rule of derivative. Once the derivatives of parameters obtained, the weight is updated as follows: the weight's output delta and input activation are multiplied to find the gradient of the weight. And then, a ratio (learning rate) of the weight's gradient is subtracted from the weight. This learning cycle is repeated until the network reaches a satisfactory validation error. More details on CNN architecture and learning algorithm can be found in \cite{Ref39, Ref40, Ref41}. 

\section{Experiments}

In order to evaluate the performance of the proposed method, the UCR time-series classification archive is used \cite{Ref42}. For having a comprehensive evaluation of the classification algorithms, this repository contains 85 time-series datasets with different characteristics i.e. the number of classes $2 \leq c \leq60$, number of training samples $16 \leq N_{tr} \leq 8926$ and time-series length $24 \leq l \leq 2709$. The datasets are obtained from varieties of different real-world applications, ranging from EEG signals to food/beverage flavors, and electric devices to phonemes signals. We have selected the same 20 datasets that the results of the other state-the-art algorithms are also reported for them (particularly considering \cite{Ref21, Ref23, Ref24} algorithms that we need to compare our results with - see Table \ref{tab1}). The training and testing sets are provided separate to make sure that the results of different algorithms of different studies are comparable. Furthermore, the error rates of "1-NN Euclidean Distance", "1-NN DTW(r) where $r$ is the percentage of time series length", and "1-NN DTW (no Warping Window)" are reported as the base-line methods. Only the results of $1-NN DTW$ is included in the table, because it obtains the lower error rates compared to the other two \cite{Ref42}. Application of RP (with the phase space dimension $m=3$, and embedding time delay $\tau = 4$) time-series to image encoding on first sample of five different datasets from the UCR archive are shown in Figure \ref{fig:RP_UCR}.

For comparison purposes, "Number of Wins" and "Average Rank" are calculated for each algorithm. Number of Wins counts number of datasets that a specific algorithm obtains the lowest error rates, while Average Rank is the mean of the error-rate ranking over entire datasets. Since some error rates are missing for some datasets for some algorithms, we normalized the measures for each algorithm. An algorithm outperforms the rest if it has a highest number of wins and lowest average rank.

\begin{table}[t]
\caption{Performance (in terms of the error rates) of the proposed method compared to the state-of-the art TSC algorithms on 20 selected data from the UCR archive.} 
\label{tab1}
\begin{center}
\small
\begin{tabular}{|l|l|l|l|l|l|l|l|l|l|l|l|l|} 
\hline
Dataset & $c$ & $N_{tr}$ & $N_{te}$ & $l$ & \scriptsize{1-NN DTW} & Shapelet & BoP & \scriptsize{SAX-VSM} & TFRP & MCNN & \scriptsize{GAF-MTF}& ours\\
\hline
50words & 50 & 450& 455 & 270 & 0.31 & 0.44 & 0.46 & - & 0.43 & \bf{0.19}  & 0.30 & 0.26\\
\hline
Adiac & 37 & 390& 391 & 176 & 0.39 & 0.51 & 0.43 & 0.38 & \bf{0.20} & 0.23 & 0.37 & 0.28\\
\hline
Beef & 5 & 30& 30 & 470 & 0.36 & 0.44 & 0.43 & \bf{0.033} & 0.36 & 0.36 & 0.23 & 0.08\\
\hline
CBF & 3 & 30& 900 & 128 & 0.003 & 0.05 & 0.01 & 0.02 & - & \bf{0.002}& 0.009 & 0.005\\
\hline
Coffee & 2 & 28& 28 & 286 & \bf{0} & 0.06 & 0.03 & \bf{0} & 0.03 & 0.036 & \bf{0} & \bf{0}\\
\hline
ECG200 & 2 & 100& 100 & 96 & 0.23 & 0.22 & 0.14 & 0.14 & 0.17 & - & 0.09 &\bf{0}\\
\hline
FaceAll & 14 & 560& 1690 & 131 & \bf{0.19} & 0.40 & 0.21 & 0.20 & 0.29 &0.23  & 0.23 & \bf{0.19}\\
\hline
Face4 & 4 & 24& 88 & 350 & 0.17 & 0.09 & 0.023 & \bf{0} & 0.21 & \bf{0} &  0.06& \bf{0}\\
\hline
Fish & 7 & 175& 175 & 463 & 0.17 & 0.19 & 0.074 & \bf{0.017} & 0.12 & 0.05 & 0.114 & 0.085\\
\hline
GunPoint & 2 & 50& 150 & 150 & 0.093 & 0.061 & 0.027 & 0.007 & 0.02 & \bf{0} & 0.08 & \bf{0}\\
\hline
Lightning2 & 2 & 60& 61 & 637 & 0.13 & 0.29 & 0.16 & 0.19 & 0.04 & 0.16 &  0.11& \bf{0}\\
\hline
Lightning7 & 7 & 70& 73 & 319 & 0.27 & 0.40 & 0.46 & 0.30 & 0.31 & \bf{0.21} &  0.26& 0.26\\
\hline
OliveOil & 4 & 30& 30 & 570 & 0.16 & 0.21 & 0.13 & \bf{0.10} & 0.13 & 0.13 &  0.2& 0.11\\
\hline
OSULeaf & 6 & 200& 242 & 427 & 0.40 & 0.35 & 0.23 & 0.107 & \bf{0.07} & 0.27 & 0.35& 0.29\\
\hline
SwedishLeaf & 15 & 500& 625 & 128 & 0.20 & 0.27 & 0.19 & 0.25 & \bf{0.04} &0.066  &0.06  & 0.06\\
\hline
SyntControl & 6 & 300& 300 & 60 & 0.007 & 0.08 & 0.03 & 0.25 & - & 0.003 &  0.007& \bf{0}\\
\hline
Trace & 4 & 100& 100 & 275 & \bf{0} & 0.002 & \bf{0} & \bf{0} & - & \bf{0} & \bf{0}& \bf{0}\\
\hline
TwoPattern & 4 & 1000& 4000 & 128 & \bf{0} & 0.11 & 0.12 & 0.004 & - & 0.002 &  0.09& 0.17\\
\hline
Wafer & 2 & 1000& 6174 & 152 & 0.02 & 0.004 & 0.003 & 0.0006 & 0.002 & 0.002 &  \bf{0}& \bf{0}\\
\hline
Yoga & 2 & 300& 3000 & 426 & 0.16 & 0.24 & 0.17 & 0.16 & 0.14 & 0.11 &  0.19& \bf{0}\\
\hline

\bf{\# wins} & - & - & - & - & 4 & 0 & 1 & 6 & 3 & 6 & 3 & \bf{10}\\
\hline
\bf{Ave.Rank} & - & - & - & - & 4.05 & 5.75 & 4.15 & 3.0 & 3.37 & 2.36 & 3.40& \bf{2.15}\\
\hline
\end{tabular}
\end{center}
\end{table}

\begin{figure}
   \begin{center}
   \begin{tabular}{c}
   \includegraphics[height=3.5cm]{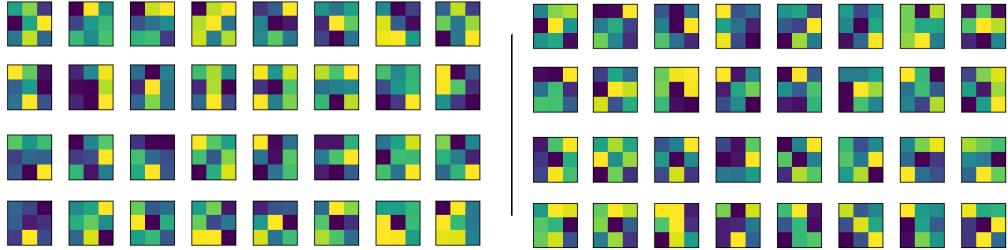}
   \end{tabular}
   \end{center}
   \caption{Visualization of kernels in the 1st (left) and 2nd (right) hidden layers from a CNN trained for "TwoPattern" data from the UCR archive.} 

  \label{fig:kernels} 
   \end{figure}

All experiments are carried out with Python\footnote{The code for the proposed model is available on: \url{http://sites.google.com/site/nimahatami/projects}}(using Keras, Theano and TensorFlow) on a PC with 2.4GHz$\times$32 CPU and 32GB memory. For training of the CNN a fixed size window as input layer is required. In the experiments 28$\times$28, 56$\times$56 and 64$\times$64 pixel input, depend on the data, are used. Both layers contain 32 feature maps with 3$\times$3 convolution (32$\times$3$\times$3), MaxPooling of size 2$\times$2 and Dropout = 0.25. The fully connected neural layer contains 128 hidden neurons and $c$ output neurons with Dropout = 0.5. A standard way to represent this architecture is \emph{C1(size)-S1-C2(size)-S2-H-O}, where $C1$ and $C2$ are number of filters in first and seconds layers, \emph{Size} denotes the kernel size, $S1$ and $S2$ are pooling size, $H$ and $O$ represent the number of hidden and output neurons in MLP. Based on this template, the proposed CNN model denoted as \emph{32(5)-2-32(5)-2-128-c}. The loss function of "categorical-crossentropy" with "adam" optimizer is used. The batch and epoch sizes are chosen from $\{5, 20\}$ and $\{50, 250, 1000, 2000\}$, respectively and based on their performance on the validation set. Since finding the optimal CNN parameters is still an open problem, we followed the rules of thumb to choose the parameters \cite{Ref41}. Visualization of kernels in the 1st and 2nd representation layers from a CNN trained for "TwoPattern" data is given in Figure \ref{fig:kernels}.

Error rate of the proposed model is given in Table \ref{tab1}. Besides comparing with $1-NN DTW$, three state-of-the art TSC algorithms are also reported: Fast-Shapelets \cite{Ref43}, Bag-of-Patterns (BoP also known as BoF or BoW) \cite{Ref44, Ref45}, and SAX-VSM \cite{Ref46}. Not all the performances are reported on the selected data. Therefore, there are some missing elements (shown with "-") in the Table. We have also compared our results with two recent deep CNN models, i.e. Multi-scale CNN(MCNN) \cite{Ref21} and GAF-MTF \cite{Ref23, Ref24}. The former performs various transformations of the time-series which extract features of different frequency and time scales embedded in a CNN framework. The latter, which has the similar pipeline to ours, first encodes time-series to images using GAF and MTF, and then applies a Tiled CNN. As shown, the proposed model obtains 10 out of 20 wins and average rank of 2.15. The second most accurate algorithm is MCNN (6 wins out of 19, and Ave.Rank = 2.36), followed by SAX-VSM (6 wins out of 19, and Ave.Rank = 3.0). The other deep CNN model that uses time-series images (GAF-MTF) ranks 5th. It highlights the fact that choosing a right time-series encoding is crucial for CNN model. Furthermore, it is observable that TFRP algorithm \cite{Ref25, Ref26}(using RP images in a traditional classification framework with different texture descriptors followed by a SVM classifier) ranks better than GAF-MTF. It can be explained that the RP introduces more discriminant and informative features than the GAF-MTF transform for TSC. Additionally, comparing the proposed method with TFRP highlights the fact that for TSC of RP images the CNN model is more effective than the traditional classification frameworks. 

\section{Conclusions}

A novel pipeline for TSC is proposed. Taking advantage of the CNN's high performance on image classification, time-series signals are first transformed into texture images (using RP) and then handled by a deep CNN model. This pipeline offers the following advantages: i) RP enables us to visualize certain aspects of the m-dimensional phase space trajectory through a 2D images, and ii) CNN automatically learns different levels of time-series features and classification jointly and in a supervised manner.  Experimental results demonstrate the superiority of the proposed pipeline. In particular, comparing with models using RP with the traditional classification framework (e.g. SIFT, Gabor and LBP features with SVM classifier \cite{Ref25, Ref26}) and other CNN-based time-series image classification (e.g. GAF-MTF images with CNN \cite{Ref23, Ref24}) demonstrates that using RP images with CNN in our proposed model obtains the better results. 

As future work, CNN architecture with more feature representation layers should be investigated for more difficult TSC tasks (preferably with more data samples available). Large datasets are needed in order to train a deeper architectures. Therefore, adopting the proposed pipeline for TSC with small sample sizes can be another interesting future direction. Exploring different ensemble learning methods \cite{Ref47} for CNN can be also interesting. We will particularly be investigating application of the output coding \cite{Ref48, Ref49, Ref50} for CNNs.  

\acknowledgments     

This research is partially supported by the French national research agency (ANR) under the PANDORE grant with reference number ANR-14-CE28-0027.



\begin{thebibliography}{1}


\bibitem{Ref1}
Wang, J., Liu, P., She, M., Nahavandi, S., Kouzani, A., Bag-of-words
representation for biomedical time series classification. Biomedical Signal
Processing and Control 8(6), 634-644, 2013.

\bibitem{Ref2} 
Hatami, N., Chira, C., Classifiers with a reject option for early time-series classification. IEEE Sym. on Computational Intelligence and Ensemble Learning (CIEL), 9-16, 2013.

\bibitem{Ref3}
Wang, Z., Oates, T., Pooling sax-bop approaches with boosting to classify
multivariate synchronous physiological time series data, FLAIRS Conference, 335-341, 2015.

\bibitem{Ref4} 
Jeong, Y., Jeong, M., Omitaomu, O., Weighted dynamic time warping
for time series classification, Pattern Recognition 44(9), 2231-2240, 2011.

\bibitem{Ref5}
Xing, Z., Pei, J., Yu, P., Early prediction on time series: A nearest
neighbor approach, International Joint Conference on Artificial Intelligence (IJCAI), 1297-1302, 2011.

\bibitem{Ref6} 
Eads, D., Glocer, K., Perkins, S., Theiler, J., Grammar-guided feature extraction for time series classification. Proceedings of the Conference on Neural Information Processing Systems (NIPS’05), 2005.

\bibitem{Ref7} 
Nanopoulos, A., Alcock, R., Manolopoulos, Y., Feature-based classification of time-series data, International Journal of Computer Research 10, 49-61, 2001.

\bibitem{Ref8} 
Rodriguez, J., Alonso, C., Interval and dynamic time warping-based decision trees, ACM Symposium On Applied Computing, 548-552, 2004.

\bibitem{Ref9} 
Krizhevsky, A. Sutskever, I. Hinton, GE. Classification with deep convolutional neural networks, Proceedings of the Conference on Neural Information Processing Systems (NIPS’12), 1097-1105, 2012.

\bibitem{Ref10} 
Simonyan, K. Zisserman, A. Very Deep Convolutional Networks for Large-Scale Image Recognition, arXiv preprint arXiv:1409.1556, 2014.

\bibitem{Ref11} 
Karpathy, A. Toderici, G. Shetty, S. Leung, T. Sukthankar, R. Fei-Fei, L. Large-scale video classification with convolutional neural networks, Computer Vision and Pattern Recognition (CVPR) Conference, 2014.


\bibitem{Ref12} 
Deng et al., Recent advances in deep learning for speech research at Microsoft, International Conference on Acoustics, Speech, and Signal Processing (ICASSP), 2013.


\bibitem{Ref13}
Graves, A. Mohamed, A. Hinton, G. Speech recognition with deep recurrent neural networks, International Conference on Acoustics, Speech, and Signal Processing (ICASSP), 2013.

\bibitem{Ref14} 
Eckmann, J. Kamphorst, S. Ruelle, D., Recurrence plots of dynamical systems, EPL (Europhysics Letters) 4(9), 973, 1987.

\bibitem{Ref15}
Haselsteiner, E. Pfurtscheller, G., Using time-dependent neural networks for EEG classification, IEEE transactions on rehabilitation engineering 8 (4), 457-463, 2000.

\bibitem{Ref16} 
Lee, H. Largman, Y. Pham, P. Ng, A. Unsupervised feature learning for audio classification using convolutional deep belief networks, Proceedings of the Conference on Neural Information Processing Systems (NIPS’09), 2009.


\bibitem{Ref17}
Zheng, Y. Liu, Q. Chen, E. Ge, Y. Zhao, J. Time Series Classification Using Multi-Channels Deep Convolutional Neural Networks, Web-Age Information Management, 298-310, 2014.

\bibitem{Ref18} 
Dalto, M. Deep neural networks for time series prediction with applications in ultra-short-term wind forecasting, IEEE International Conference on Industrial Technology (ICIT), 2015.

\bibitem{Ref19} 
Abdel-Hamid, O. Deng, L. Yu, D. Exploring convolutional neural network structures and optimization techniques for speech recognition, Conference of the International Speech Communication Association (Interspeech), 3366-3370, 2013.

\bibitem{Ref20} 
Abdel-Hamid, O. Mohamed, A. Jiang, H. Penn, G. Applying convolutional neural networks concepts to hybrid NN-HMM model for speech recognition, International Conference on Acoustics, Speech, and Signal Processing (ICASSP), 2012.


\bibitem{Ref21} 
Cui, Z. Chen, W. Chen, Y. Multi-scale convolutional neural networks for time series classification, arXiv:1603.06995, 2016. 

\bibitem{Ref22}
Yang, J. Nguyen, M. San, P. Li, X. Krishnaswamy, S. Deep Convolutional Neural Networks on Multichannel Time Series for Human Activity Recognition, International Joint Conference on Artificial Intelligence (IJCAI) 2015.

\bibitem{Ref23}
Wang, Z., Oates, T., Imaging Time-Series to Improve Classification and Imputation, International Joint Conference on Artificial Intelligence (IJCAI) 3939-3945, 2015.


\bibitem{Ref24}
Wang, Z., Oates, T. Encoding time series as images for visual inspection and classification using tiled convolutional neural networks, Association for the Advancement of Artificial Intelligence (AAAI) conference, 2015.

\bibitem{Ref25}
Souza, V. Silva, D. Batista, G. Extracting Texture Features for Time Series Classification, 1425-1430, International Conference on Pattern Recognition (ICPR), 2014.

\bibitem{Ref26}
Souza, V. Silva, D. Batista, G. Time Series Classification Using Compression Distance of Recurrence Plots, 687-696, IEEE International Conference on Data Mining (ICDM), 2013.


\bibitem{Ref27} 
Marwan, N., Romano, M.C., Thiel, M., Kurths, J., Recurrence plots for
the analysis of complex systems, Physics Reports 438(5-6), 237-329, 2007.


\bibitem{Ref38} 
Krizhevsky, A. Sutskever, I. Hinton, GE. Imagenet classification with deep convolutional neural networks, Proceedings of the Conference on Neural Information Processing Systems (NIPS’12), 1097-1105, 2012.



\bibitem{Ref39} 
LeCun, Y. Bottou, L. Bengio, Y. Haffner, P. Gradient-based learning applied to document recognition, Proceedings of the IEEE 86 (11), 2278-2324, 1998.

\bibitem{Ref40} 
Bouvrie, J. Notes on convolutional neural networks, 2006.

\bibitem{Ref41} 
LeCun, Y. Bottou, L. Orr, G. Müller, K. Efficient backprop, Neural networks: Tricks of the trade, 9-48, 2012.

\bibitem{Ref42} 
Chen, Y., Keogh, E., Hu, B., Begum, N., Bagnall, A., Mueen, A., Batista, G., The ucr time series classification archive, URL \url{www.cs.ucr.edu/~eamonn/time_series_data/}, 2015.

\bibitem{Ref43} 
Rakthanmanon, T. Keogh, E. Fast shapelets: A scalable algorithm for discovering time series shapelets, SIAM International Conference on Data Mining, 668-676, 2013.

\bibitem{Ref44} 
Lin, J., Khade, R., Li, Y., Rotation-invariant similarity in time series using bag-of-patterns representation, Journal of Intelligent Information Systems 39(2), 287-315, 2012.

\bibitem{Ref45} 
Oates, T. et al. Exploiting representational diversity for time series classification, International Conference On Machine Learning And Applications (ICMLA), 2012. 

\bibitem{Ref46}
Senin, P., Malinchik, S., Sax-vsm: Interpretable time series classification
using sax and vector space model, 1175-1180, International Conference on Data Mining (ICDM), 2013.

\bibitem{Ref47} 
Hatami, N. Some proposal for combining ensemble classifiers, PhD thesis University of Cagliari, 2012.

\bibitem{Ref48} 
Hatami, N., Thinned-ecoc ensemble based on sequential code shrinking,
Expert Systems with Applications 39(1), 936-947, 2012.

\bibitem{Ref49} 
Hatami, N., Ebrahimpour, R., Ghaderi, R., Ecoc-based training of neural
networks for face recognition, IEEE Conference on Cybernetics and Intelligent
Systems , 450-454, 2008.

\bibitem{Ref50} 
Armano, G. Chira, C. Hatami, N. Error-Correcting Output Codes for Multi-Label Text Categorization, 26-37, Italian Information Retrieval Workshop (IIR), 2012.








\end{thebibliography}
\end{document}